\documentclass{article}
\usepackage{spconf,amsmath,graphicx}
\usepackage{amsmath}
\usepackage{cases}

\title{Spatial morphing kernel regression for feature interpolation}
 \name{Xueqing Deng, Yi Zhu and Shawn Newsam}
	\address{University of California, Merced\\
	Electrical Engineering and Computer Science\\
	5200 N Lake Rd, Merced, CA, United States}
\begin{document}
%
\maketitle
\begin{abstract}
In recent years, geotagged social media has become popular as a novel source for geographic knowledge discovery. Ground-level images and videos provide a different perspective than overhead imagery and can be applied to a range of applications such as land use mapping, activity detection, pollution mapping, etc. The sparse and uneven distribution of this data presents a problem, however, for generating dense maps. We therefore investigate the problem of spatially interpolating the high-dimensional features extracted from sparse social media to enable dense labeling using standard classifiers. Further, we show how prior knowledge about region boundaries can be used to improve the interpolation through spatial morphing kernel regression. We show that an interpolate-then-classify framework can produce dense maps from sparse observations but that care must be taken in choosing the interpolation method. We also show that the spatial morphing kernel improves the results.
\end{abstract}
\section{Introduction}
\label{sec:intro}
Mapping geographic phenomena on the surface of the Earth is an important scientific problem. Remote sensing is a traditional approach in which analysis is performed on overhead images from satellites and aircraft. This can produce dense maps but is limited by the overhead view. For example, one cannot see inside buildings.

The widespread availability of geotagged social media has enabled novel approaches to geographic discovery. In particular, ``proximate sensing'' \cite{leung2010proximate} using ground-level images and videos available at sharing sites like Flickr and YouTube provides a different perspective from remote sensing, one that can see inside buildings and detect phenomena not observable from above. Proximate sensing has been applied to map land use classes \cite{zhu2015land}, public sentiment \cite{zhu2016spatio}, human activity \cite{Zhu2017Activity}, air pollution \cite{li2015using}, and natural events \cite{wang2016tracking}, among other things.

\begin{figure}
\centering
\includegraphics[width=4cm]{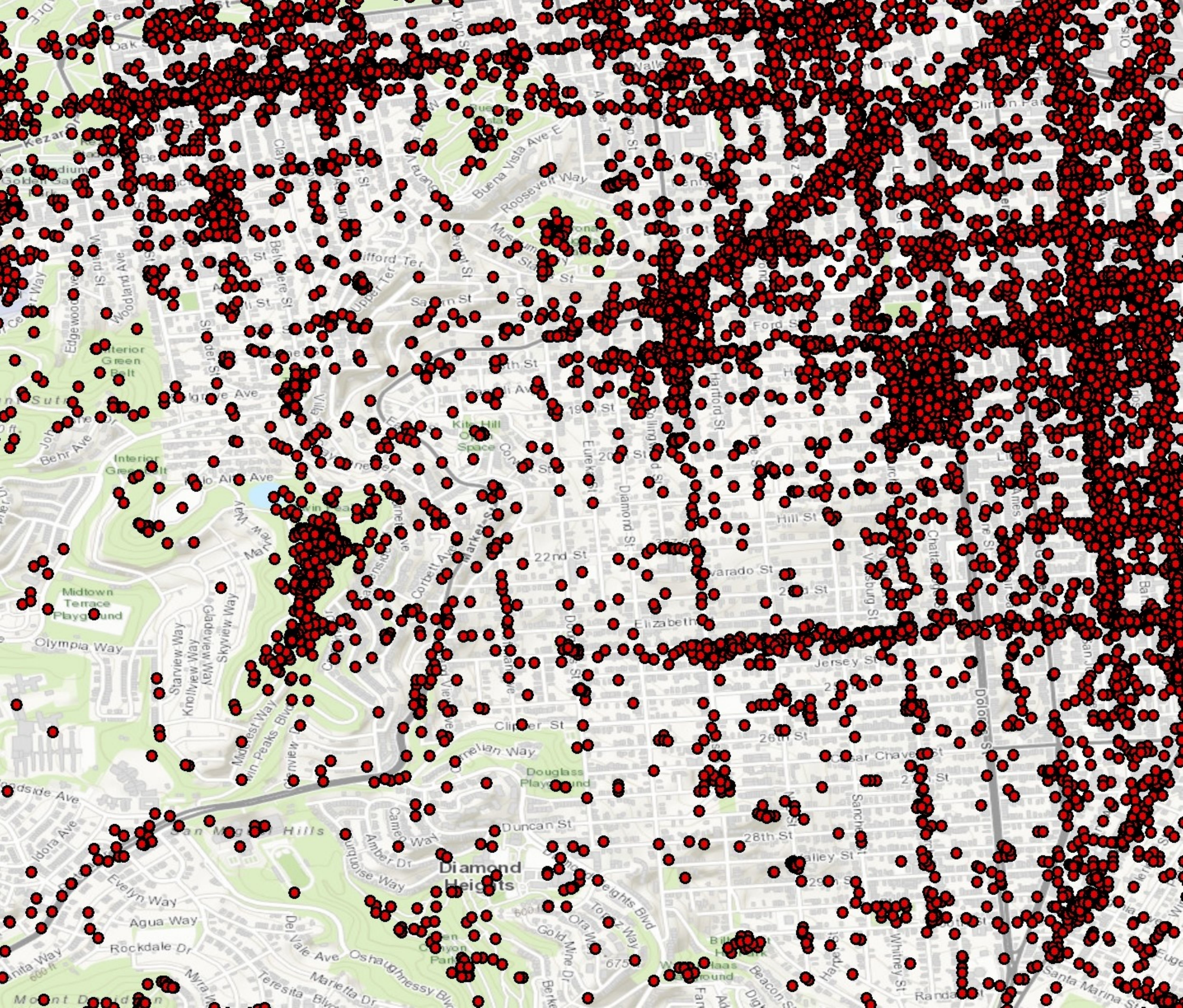}
\caption{Distribution of Flickr images in San Francisco. While these images can be used to map geographic phenomena such as land use, the resulting maps are sparse and uneven. We therefore investigate methods to interpolate the high-dimensional image features before performing classification.}
\label{fig:flickr_sf}
\end{figure}

A fundamental challenge in using geotagged social media to create dense maps is its sparse and uneven spatial distribution. For example, figure \ref{fig:flickr_sf} shows the spatial distribution of Flickr images for a region of San Francisco. Even if one was able to use these images to accurately label land use, for example, the resulting map would itself be sparse and uneven. 

We therefore investigate an alternate approach in which the high-dimensional features extracted from the geotagged social media are spatially interpolated before classification is performed. To our knowledge, there has been very little work on this interpolate-then-classify problem. Workman et al. in \cite{workman2017unified} spatially interpolate features extracted from Google Street View images to match the spatial density of features extracted from overhead imagery. But, they do not investigate how best to do the interpolation. Our work in this paper performs an in-depth evaluation of the interpolate-then-classify problem using synthetic as well as real datasets.

\begin{figure*}
\centering
\includegraphics[width=15cm]{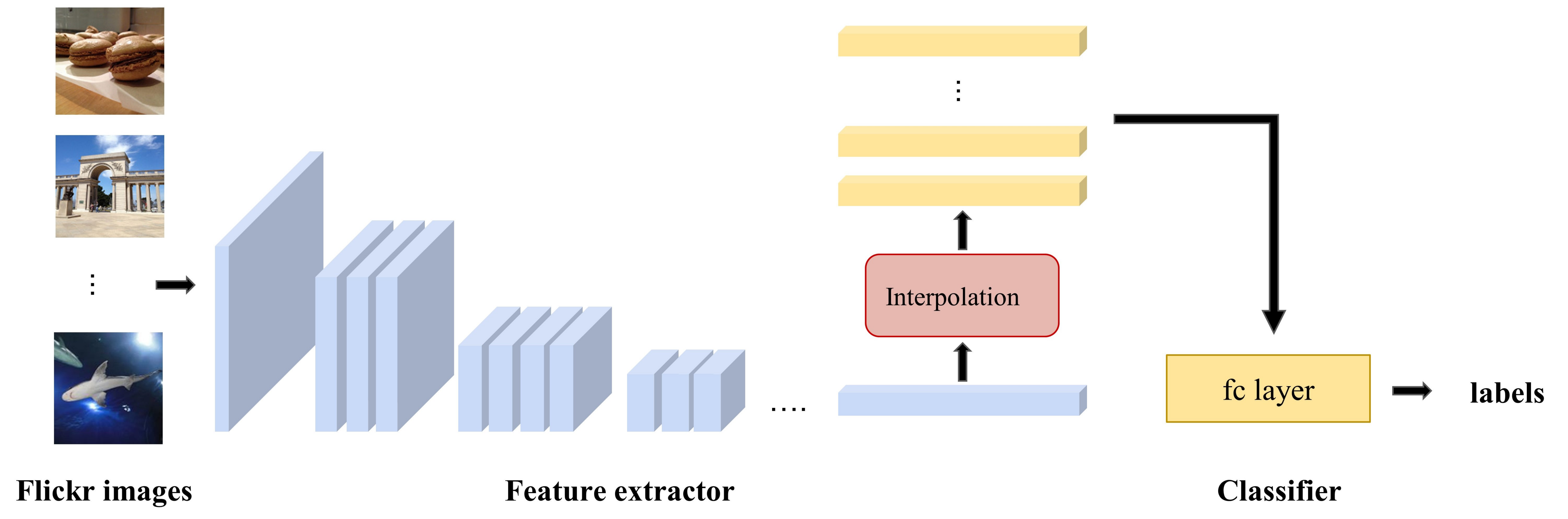}
\caption{Our proposed interpolate-then-classify framework. The convolutional layers of a CNN (blue) are used to perform feature extraction on sparsely located ground-level images. We investigate various interpolation methods (red) including ones that incorporate prior knowledge of spatial heterogeneity. The fc layer of the CNN (yellow) is then used to perform dense classification.}
\label{fig:framework}
\end{figure*}

We also investigate how to use prior knowledge about spatial heterogeneity to modulate the interpolation. We take inspiration from \cite{pozdnoukhov2010spatial} which proposes a novel kernel that incorporates prior knowledge on spatial similarities, discontinuities, and physical and administrative boundaries to spatially interpolate a continuous variable. For example, \cite{pozdnoukhov2010spatial} shows that knowledge of building boundaries can improve the interpolation of temperature as the indoor and outdoor temperature can be quite different. However, \cite{pozdnoukhov2010spatial} only interpolates a single continuous variable and the interpolation is the final result--no classification is performed. We instead interpolate high-dimensional features extracted by convolutional neural networks (CNNs) with the goal of performing dense classification. The prior knowledge is incorporated through a graph Laplacian. We consider two types of graph Laplacians, one constructed using a mesh grid and another constructed using the sparse feature locations themselves.

To summarize the salient aspects of our work, we investigate the novel problem of spatially interpolating high dimensional features for dense geographic classification. We incorporate prior knowledge of spatial heterogeneity through spatial morphing kernels. And, we show results using synthetic as well as real data for mapping land use.

\vspace{-5pt}
\section{Methodology}
Our framework consists of three steps as shown in figure \ref{fig:framework}: feature extraction, feature interpolation, and dense classification. We use a pre-trained CNN without the final fully connected (fc) layers to perform the feature extraction. We investigate various interpolation methods including ones that incorporate prior knowledge of spatial heterogeneity. Finally, the fc layers of the CNN are used to classify the densely interpolated features.
\vspace{-5pt}
\subsection{Convolutional Neural Network}
Our CNN is a ResNet-101 \cite{he2016deep} model that has been trained to label ground-level images as depicting one of 45 different land use classes. (Please see \cite{zhu2018arxiv} for more details on this model.) We separate the network into two parts: 1) a feature extractor consisting of the convolutional layers that outputs a 2,048 dimensional feature vector, and 2) a classifier consisting of the fc layer.
\vspace{-5pt}
\subsection{Interpolation}
Our interpolation problem is defined as follows. Suppose we have a sparse set of $n$ image locations \textit{\textbf{S}}=\{$s_{1},s_{2},...,s_{n}$\} from which we have extracted high-dimensional features $f(s_{i})$. Our goal is to use these features and their locations to estimate the feature at a novel location $f(l)$. We can then create a dense feature map by densely sampling the locations $l$. We now describe the different interpolation methods we consider.

\subsubsection{Inverse Distant Weighting}
Inverse distance weighting (IDW) \cite{Shepard1968} is a commonly used approach to interpolate a spatially smooth surface. IDW assumes that locations that are close to one another are more alike than those that are far apart. IDW interpolation is computed as 

\begin{table*}[ht]
\centering
\caption{Quantitative results on the synthetic data. The rows indicate the number of images per region. The results under the columns IDW, Gaussian, SMSK, and SMMK is the average mIoU over 20 trials. The noise column shows the percentage of spurious classes introduced by the interpolation method to the left.}
\begin{tabular}{|c|c|c|c|c|c|c|c|c|}
\hline
Method& IDW &noise (\%) & Gaussian & noise (\%)&SMSK&noise (\%) &SMMK& noise (\%)\\
\hline
1 & 45.5	& 13.4 &	72.6 &	1.9	& \textbf{73.2} &	0.5 &	70.2	& 2.9\\
2 & 56.8	& 3.1	& 74.6	& 1.5	& \textbf{77.2}	&0.1	&75.5 &1.5\\
3 & 61.1	& 4.8	&77.1	& 1.3	&77.8	&0.2	& \textbf{79.8} &	1.5\\
5 &68.1	&17.8	&81.3	& 3.7	&81.8	& 0.4	& \textbf{83.9}	& 2.3 \\
10&80.5	&4.9&	83.6&	1.9&	84.7&	0.0&	\textbf{87.4}&	1.8\\

\hline
\end{tabular}
\label{tab:quan_toy}
\end{table*}

\begin{equation}
 f (\textit{l})=\begin{cases}
     \sum_{i=1}^{N}w_{i}(\textit{l})f(s_{i}), & \text{if $d(l,s_{i})\neq 0$ for all $i$}.\\
    f(s_{i}), & \text{if $d(l,s_{i})=0$ for some $i$}.
  \end{cases}
\end{equation}
where $N$ is the number of locations used to perform the interpolation, and $w_{i}(l)={1}/{d(l,s_{i})}$ is the weight given to feature of the $i$th location. $d(l,s_{i})$ is commonly computed as the Euclidean distance between locations $l$ and $s_{i}$ in 2D geographic space.

\subsubsection{Kernel Regression}
We also interpolate the features using Nadaraya-Waston kernel regression as is done in \cite{workman2017unified}. This interpolation is computed as

\begin{equation}
f(l)=\frac{\sum_{i=1}^{N}w_{i}(l)f(s_{i})}{\sum_{j=1}^{N}w_{j}(l)}
\end{equation}
where $w_{i}(l)=\it{k}(\bf{x},\bf{x^{'}})$ is a kernel based on the locations $\bf{x}$ and $\bf x^{'}$. In our case, $\bf{x}$ is the 2D location $l$ and $\bf x^{'}$ is the 2D location $s_i$. We consider a standard Gaussian kernel as well as a spatial morphing kernel inspired by \cite{pozdnoukhov2010spatial}.

\noindent {\bf Gaussian kernel:} \textit{k} ($\bf{x},\bf{x^{'}}$) = $\exp(-d(l,s_{i};\Sigma)^{2})$ where $d(l,s_{i};\Sigma)$ is the normalized Euclidean distance. The diagonal covariance matrix $\Sigma$ controls the kernel bandwidth. For our 2D case, $\Sigma$ contains two non-zero values which are learned during a training phase.

\noindent {\bf Spatial morphing kernel:} \cite{pozdnoukhov2010spatial} describes how a graph Laplacian based on spatial adjacency/connectivity can incorporate prior spatial knowledge into the regression kernel. First, an adjacency matrix \textbf{W} is computed in which $w_{ij} = 1$ if locations $i$ and $j$ are connected and $w_{ij} = 0$ otherwise. This can encode, for example, the connectivity between locations given building boundaries, etc. The graph Laplacian is then computed as $\bf{L=D-W}$ where \textbf{D} is the diagonal node degree matrix in which $d_{ii}=\sum_{j}w_{ij}$. The spatial morphing kernel is then computed as
\begin{equation}
\tilde{k}(\bf{x},\bf{x^{'}})=\it{k}(\bf{x},\bf{x^{'}})-k^{T}_{x}(I+\gamma LK)^{-1}\gamma L k_{x^{'}}
\end{equation}
where $\bf{I}$ is the identity matrix, $\bf{K}$=$\{k(x_{i},x_{j})\}_{i,j=1,...,N}$ is the kernel matrix for all data samples, and $\bf{k}_{x}$ and $\bf k_{x^{'}}$ are the vectors [$k(x,x_1),...,k(x,x_{N})$] and [$k(x^{'},x_1)$,...,$k(x^{'},x_{N})$]. ($k(\cdot,\cdot)$ is  the Gaussian kernel above.) The hyper parameter $\gamma$ controls how much the kernel is spatially morphed based on the prior knowledge.

We compute the adjacency matrix $\bf{W}$ in two ways. First, similar to \cite{pozdnoukhov2010spatial}, a dense mesh grid is laid over the study area and the locations of this grid form the nodes of the graph. We refer to this as the spatial morphing mesh kernel (SMMK). Second, the nodes of the graph are the locations of our observed features (our images).  We refer to this as the spatial morphing sample kernel (SMSK). In either case, we set $w_{ij} = 1$ if locations $i$ and $j$ are in the same region and $w_{ij} = 0$ otherwise.

\section{Experiments}
We focus on the problem of dense land use mapping from sparse ground level images. We conduct experiments using synthetic as well as real data. We know the ground truth land use map for the synthetic data which allows us to perform quantitative evaluation. The Gaussian kernel bandwidth is tuned using leave-one-out cross validation, and we set $\gamma=100$ for the spatial morphing kernels.
 
 \begin{figure*}[ht]
\centering
\includegraphics[width=\textwidth]{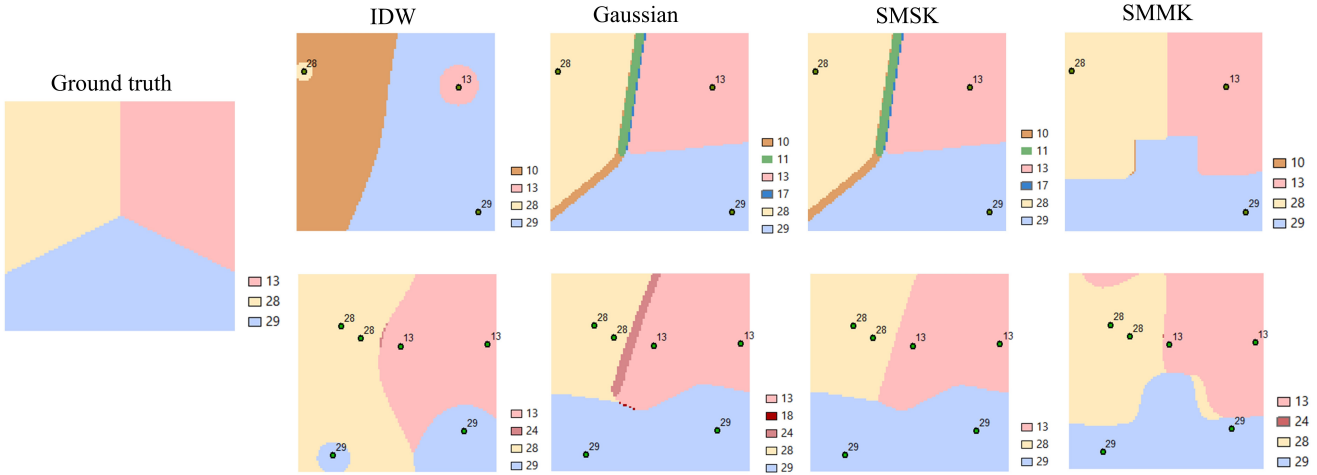}
\caption{Selected results on the synthetic data. The left is the ground truth map we are trying to estimate. The top row on the right are the results for one image per ground truth region. The bottom row is for two images.}
\label{fig:simu_toy}
\end{figure*}
\begin{figure*}[ht]
\centering
\includegraphics[width=\textwidth]{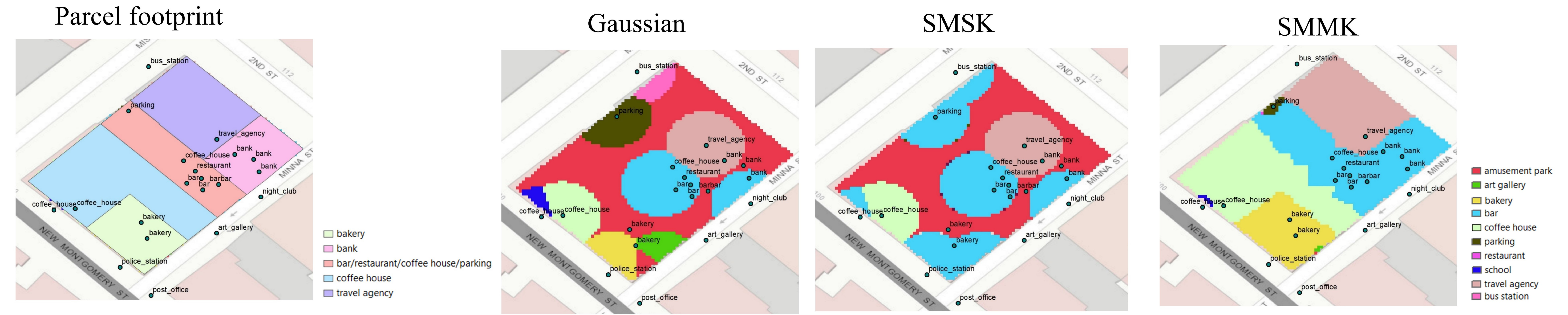}
\caption{Selected results on the real world data from San Francisco.}
\label{fig:sf_results}
\end{figure*}

\subsection{Synthetic Data}
We partition a 100m $\times$ 100m piece of land into three regions with different land use classes as shown on the left in figure \ref{fig:simu_toy}. This is considered the ground truth. Our goal is to generate this map at the 1m $\times$ 1m scale from sparsely located images. We then randomly pick three (real) ground level images from our San Francisco dataset with different land use labels (as assigned by our CNN). We then place these images at random locations in the ground truth regions. We consider different densities of images: 1, 2, 3, 5, and 10 per region. See figure \ref{fig:simu_toy} for example configurations.

We then perform feature interpolation over the entire 100 $\times$ 100 area. We compute one feature per 100 $\times$ 100 location which is then used to classify the location. The resulting map is compared to the ground truth by computing the mean Intersection over Union (mIoU) \cite{long2015fully} between the predicted regions and the true regions using the 100 $\times$ 100 grid. We also compute the percentage of locations that are assigned a class other than the three in the ground truth. This gives an indication of how stable the interpolation is with respect to the classifier. We call this value the noise percentage. We compute the average mIoU and noise percentage over 20 trials.

The graph Laplacian for the spatial morphing kernels is computed using the known locations of the region boundaries.
 
\subsection{Real Dataset: San Francisco}
We create a land use map of San Francisco using ground level images from Flickr. We download land parcel footprints from the City of San Francisco website. These footprints are used to construct the graph Laplacians for the spatial morphing kernels.
\vspace{-10pt}
\section{Results and discussion}
Figure \ref{fig:simu_toy} shows the results for the synthetic data. The top row corresponds to one image per region and the bottom row corresponds to two images per region. IDW interpolation is seen to have difficulty especially when the images are very sparse. For example, it actually introduces a completely new class (10) in the one image case. Gaussian kernel regression is see to perform much better. This demonstrates the importance of the interpolation method in our framework.

The spatial morphing kernels produce maps with more accurate boundaries and fewer spurious classes. As shown in the bottom row of figure \ref{fig:simu_toy}, for the two image case, Gaussian kernel regression introduces two spurious classes while SMSK do not introduce any. This shows the SMSK interpolation is more stable with respect to the classifier. 

Table \ref{tab:quan_toy} shows the qualitative results for the synthetic data. SMSK and SMMK outperform the interpolation methods that do not incorporate prior spatial knowledge. SMSK does better than SMMK for very sparse image configurations while the opposite is true for more dense configurations. We will investigate this further in future work.

Figure \ref{fig:sf_results} shows the results for the real data of San Francisco. On the left is shown the land parcels with the labels that have been assigned to the Flickr images by our classifier. (This is not the ground truth--in general, we do not know the ground truth for the real data.) On the right are the maps produced using the different interpolation methods. Here, SMMK clearly produces the best results. Its map has much more accurate boundaries than the other approaches. It is also less affected by images that fall outside the parcels.

See the supplementary materials for more details on the synthetic and real data.
\vspace{-10pt}
\section{Conclusion}
We investigated the problem of spatially interpolating high-dimensional features extracted from sparse ground-level images for dense mapping. We compared different interpolation methods including ones which incorporate prior knowledge of spatial heterogeneity. We evaluated the methods on synthetic as well as real data. We observed that the choice of interpolation method is important. We also showed that the methods that incorporate spatial knowledge result in more accurate regions.

In future work, we plan to have tighter integration between the CNN and the interpolation. One possible way to do this is train the CNN to be more robust with respect to interpolation between classes that frequently occur next to each other geographically.


\bibliographystyle{IEEEbib}
\bibliography{refs}

\end{document}